%% file: bmvc_arxiv.tex
\documentclass{bmvc2k}

\usepackage{multirow}
\usepackage{blindtext}
\usepackage{placeins}
\usepackage{adjustbox}
\usepackage{xfrac}
\usepackage{enumitem}
\usepackage{subcaption}
\usepackage{array}
\usepackage{rotating}
\usepackage{multicol}
\usepackage{chngcntr}
\usepackage{bm}
\usepackage{booktabs}
\usepackage{comment}
\usepackage{lipsum}
\usepackage{wrapfig}
\usepackage{etoolbox}
\BeforeBeginEnvironment{wrapfigure}{\setlength{\intextsep}{0pt}}

\usepackage{ifthen}
\newboolean{ack}
\newboolean{combined}
\setboolean{ack}{true} 
\setboolean{combined}{true}

\definecolor{Blue}{rgb}{.1,.1,.8}
\definecolor{Red}{rgb}{1,0,0}
\definecolor{Green}{rgb}{.1,.8,.1}

\title{Self-Supervised Training Enhances Online Continual Learning}

\addauthor{Jhair~Gallardo}{gg4099@rit.edu}{1}
\addauthor{Tyler~L.~Hayes}{tlh6792@rit.edu}{1}
\addauthor{Christopher~Kanan}{kanan@rit.edu}{1,2,3}

\addinstitution{
 Rochester Institute of Technology\\
 New York, USA
}
\addinstitution{
 Paige\\
 New York, USA
}
\addinstitution{
 Cornell Tech\\
 New York, USA
}

\runninghead{Gallardo, Hayes, Kanan}{Self-supervised training enhances OCL}

\newcommand{\beginsupplement}{%
        \setcounter{table}{0}
        \renewcommand{\thetable}{S\arabic{table}}%
        \setcounter{figure}{0}
        \renewcommand{\thefigure}{S\arabic{figure}}%
        
        \setcounter{section}{0}
        \renewcommand{\thesection}{S\arabic{section}}%
        
        \setcounter{page}{1}
        \renewcommand{\thepage}{S\arabic{page}}%
}

\begin{document}

\maketitle

\begin{abstract}
In continual learning, a system must incrementally learn from a non-stationary data stream without catastrophic forgetting. Recently, multiple methods have been devised for incrementally learning classes on large-scale image classification tasks, such as ImageNet. State-of-the-art continual learning methods use an initial supervised pre-training phase, in which the first 10\% - 50\%  of the classes in a dataset are used to learn representations in an offline manner before continual learning of new classes begins. We hypothesize that self-supervised pre-training could yield features that generalize better than supervised learning, especially when the number of samples used for pre-training is small. We test this hypothesis using the self-supervised MoCo-V2, Barlow Twins, and SwAV algorithms. On ImageNet, we find that these methods outperform supervised pre-training considerably for online continual learning, and the gains are larger when fewer samples are available. Our findings are consistent across three online continual learning algorithms. Our best system achieves a 14.95\% relative increase in top-1 accuracy on class incremental ImageNet over the prior state of the art for online continual learning.
\end{abstract}

\section{Introduction}
\label{sec:intro}

Conventional convolutional neural networks (CNNs) are trained offline and then evaluated. When new training data is acquired, the CNN is re-trained from scratch. This can be wasteful for both storage and compute.
Ideally, the CNN would be updated on only new samples, which is known as continual learning. The longstanding challenge has been that catastrophic forgetting~\cite{mccloskey1989catastrophic} occurs in conventional CNNs when updating on only new samples, especially when the data stream is non-stationary (e.g., when classes are learned incrementally)~\cite{kemker2017measuring,belouadah2020comprehensive,de2019continual}. Recently, continual learning methods have scaled to incremental class learning on full-resolution images in the 1000 category ImageNet dataset with only a small gap between them and offline systems~\cite{hayes2019remind,hayes2019lifelong,rebuffi2017icarl,castro2018end,zhang2020class,hou2019learning,wu2019large}. These methods use an initial \emph{supervised} pre-training period on the first 10--50\% of the classes before continual learning begins.

Performance of these systems depends on the amount of data used for supervised pre-training~\cite{hayes2019lifelong}. We hypothesized that self-supervised pre-training would be more effective, especially when less data is used. Because supervised learning only requires features that discriminate among the classes in the pre-training set, it may not produce optimal representations for unseen categories. In contrast, self-supervised learning promotes the acquisition of category agnostic feature representations~\cite{ericsson2021well,caron2020unsupervised}. Using these features could help close the generalization gap between CNNs trained offline and those trained in a continual manner. Here, we test this hypothesis using three self-supervised learning methods.

\begin{table}[t]
	\begin{minipage}{0.485\linewidth}
    \begin{center}
      \includegraphics[width=\linewidth]{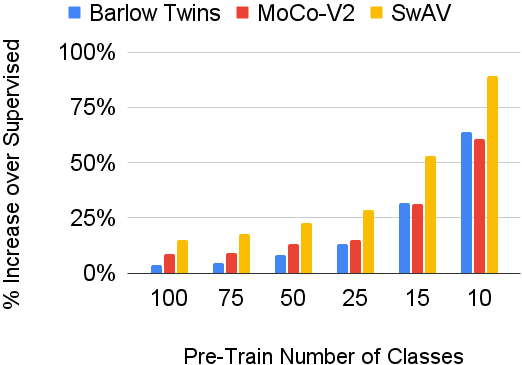}
    \end{center}
		\captionof{figure}{In this study, we found that self-supervised pre-training was more effective than supervised pre-training on ImageNet when less data was used. This figure shows the percentage increase in final top-1 accuracy on ImageNet of three self-supervised methods, MoCo-V2~\cite{chen2020improved}, Barlow Twins~\cite{zbontar2021barlow}, and SwAV~\cite{caron2020unsupervised}, compared to supervised pre-training when paired with REMIND~\cite{hayes2019remind} as a function of the number of ImageNet classes used for pre-training.}
		\label{fig:main}
	\end{minipage}\hfill
	\begin{minipage}{0.485\linewidth}
    \begin{center}
      \includegraphics[width=\linewidth]{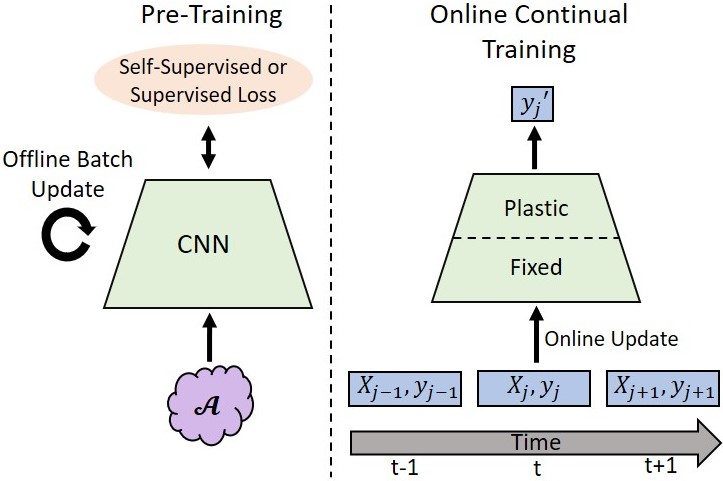}
    \end{center}
		\captionof{figure}{During pre-training (left), a CNN is trained offline on a (possibly labeled) dataset, $\mathcal{A}$. These initialized weights are copied into the continual learner's CNN, where a portion of the network is frozen and the rest is updated continually. In online learning (right), the learner receives a single labeled sample $\left(X_{j}, y_{j}\right)$ at each time-step $t$. The continual learner first learns all samples from the pre-train dataset, $\mathcal{A}$, before learning samples from dataset $\mathcal{B}$.}
		\label{fig:methods}
	\end{minipage}
	\vspace{-1em}
\end{table}

\textbf{This paper makes the following contributions:} \textbf{(1)} We compare the discriminative power of supervised features to self-supervised features learned with MoCo-V2~\cite{chen2020improved}, Barlow Twins~\cite{zbontar2021barlow}, and SwAV~\cite{caron2020unsupervised} as a function of the amount of training data used during pre-training. In offline linear evaluation experiments, we find that self-supervised features generalize better to ImageNet categories omitted from pre-training, with the gap being larger when less data is used. \textbf{(2)} We further study the ability of supervised and self-supervised features on datasets that were not used for pre-training. \textbf{(3)} We study the effectiveness of self-supervised features in three systems for online continual learning of additional categories in ImageNet. Across these algorithms, we found self-supervised features worked better when fewer categories were used for pre-training compared to supervised pre-training (see Fig.~\ref{fig:main}). \textbf{(4)} We set a new state of the art for online continual learning of ImageNet, where data is ordered by category, with a relative increase of 14.95\% over prior work~\cite{hayes2019remind}.

\section{Problem Formulation}
\label{sec:problem_formulation}

We study the common continual learning paradigm in which pre-training precedes continual learning~\cite{hayes2019remind,hayes2019lifelong,rebuffi2017icarl,castro2018end,zhang2020class,hou2019learning,kemker2017fearnet,kemker2017measuring,wu2019large,belouadah2018deesil}. Formally, given a pre-training dataset $\{\left(X_i , y_i\right) \} \in \mathcal{A}$, with $M$ images $X_i$ and their corresponding labels $y_i \in \mathcal{Y}$, a set of parameters $\theta$ are learned for a CNN using $\mathcal{A}$ in an offline manner, i.e., the learner can shuffle and loop over data. Here, we compare learning $\theta$ with supervised versus self-supervised approaches.

Continual learning begins after pre-training by first initializing the CNN parameters $\psi$ to be continually updated to $\psi \leftarrow \theta$. The continual learning dataset $\{\left(X_j , y_j \right)\} \in \mathcal{B}$, with images $X_j$ and their corresponding labels $y_j \in \mathcal{K}$ is then incrementally presented to the learner, which cannot loop over $\mathcal{B}$. We study the most general form of the problem, online continual learning, where examples are presented one-by-one to the learner and cannot be revisited without using auxiliary storage, which is kept at a fixed capacity. In incremental class learning, examples are presented in order of their category, and categories during pre-training and continual learning are disjoint, i.e, $\mathcal{Y} \cap \mathcal{K} = \emptyset$. Our training protocol is depicted in Fig.~\ref{fig:methods}.

We focus on online continual learning because these systems are more general and can learn from data presented in any order~\cite{hayes2019remind,hayes2019lifelong,hayes2019memory}, while most incremental batch learning implementations are bespoke to incremental class learning and require significant algorithmic changes for other orderings. In incremental batch learning, a system queues up examples in memory until it acquires a batch, which is typically about 100000 examples (100 classes) in papers using ImageNet~\cite{rebuffi2017icarl,castro2018end,wu2019large}. Systems loop over the batch and then purge it from memory\footnote{With the exception of those samples cached in an auxiliary storage buffer, if one is used.}, and the next batch is acquired. The online formulation is more general, i.e., the batch size is one sample, it closely matches real-world applications, systems train $5 - 130\times$ faster~\cite{hayes2019remind,hayes2019lifelong}, and recent work has shown it works nearly as well as batch learning~\cite{hayes2019remind}.

\section{Related Work}
\label{sec:related_work}

\subsection{Continual Learning}
\label{subsec:rw_continual_learning}

Deep neural networks suffer from catastrophic forgetting~\cite{mccloskey1989catastrophic} when they are incrementally updated on non-stationary data streams. Catastrophic forgetting occurs when past representations are overwritten with new ones, causing a drop in performance on past data. There are three approaches to mitigating catastrophic forgetting~\cite{parisi2019continual}: 1) increasing model capacity to include new representations~\cite{wang2017growing,rusu2016progressive,roy2020tree,aljundi2017expert,rosenfeld2018incremental,rebuffi2018efficient,mallya2018packnet,mallya2018piggyback,tao2020few}, 2) regularizing parameter updates such that parameters do not deviate too much from their past values~\cite{aljundi2018memory,chaudhry2018riemannian,chaudhry2019efficient,dhar2019learning,kirkpatrick2017,li2017learning,lopez2017gradient,ritter2018online,serra2018overcoming,zenke2017continual}, and 3) replay (or rehearsal) models that cache previous data in an auxiliary memory buffer and mix it with new data to fine-tune the network~\cite{hayes2021replay,rebuffi2017icarl,castro2018end,hou2019learning,wu2019large,hayes2019remind,douillard2020podnet,belouadah2019il2m,kemker2017fearnet,tao2020topology}. While network expansion and regularization methods have been popular, they do not easily scale to large datasets (see \cite{belouadah2020comprehensive} for a review). Moreover, many of these methods require additional information such as task labels and task boundaries,
or they perform poorly~\cite{kemker2017measuring,hayes2019remind,hayes2019lifelong}. Our experiments use a single shared classifier during online learning, where task labels are unknown to models during training and evaluation. This setting is more realistic since task information isn't always available.

Recently, methods that use replay have demonstrated success for large-scale continual learning of the ImageNet dataset~\cite{rebuffi2017icarl,castro2018end,hou2019learning,wu2019large,hayes2019remind,douillard2020podnet,belouadah2019il2m,tao2020topology}. All of these models follow a similar training paradigm where they are first pre-trained on a subset of 100~\cite{rebuffi2017icarl,castro2018end,wu2019large,hayes2019remind} or 500~\cite{hou2019learning,douillard2020podnet,tao2020topology} ImageNet classes in an offline setting, where they can shuffle and loop over the data in batches. After pre-training, these methods are updated on the remaining ImageNet classes. Although the algorithms differ fundamentally, they all perform pre-training using a supervised cross-entropy loss and a fixed pre-train number of classes. Only \cite{hayes2019lifelong} has investigated the impact of the size of the supervised pre-training set on a continual learner's performance. However, we think this is an important aspect that should be investigated further as smaller pre-training sizes require less compute and less data. 

\subsection{Self-Supervised Learning}
\label{subsec:rw_selfsupervised_learning}

Self-supervised learning techniques have recently gained popularity because they now rival supervised learning techniques~\cite{caron2020unsupervised} without requiring labeled data, which can be expensive and difficult to obtain. Specifically, self-supervised learning methods use pretext tasks to learn visual features, where the network provides its own supervision during training.
Different pretext tasks have been proposed including the prediction of image colorization~\cite{zhang2016colorful}, image rotation~\cite{gidaris2018unsupervised}, and several others~\cite{noroozi2016unsupervised,dosovitskiy2014discriminative,doersch2015unsupervised,caron2018deep}. Recent works use contrastive learning~\cite{hadsell2006dimensionality,oord2018representation}, where the model learns which datapoints are similar or different based on feature similarity, and they have even surpassed supervised networks on many downstream tasks~\cite{he2020momentum,chen2020simple,chen2020improved,chen2020big,li2020prototypical,grill2020bootstrap,caron2020unsupervised}.
Popular contrastive learning methods include MoCo~\cite{he2020momentum}, SimCLR~\cite{chen2020simple}, MoCo-V2~\cite{chen2020big}, SimCLR-V2~\cite{chen2020big}, and SwAV~\cite{caron2020unsupervised}.
SwAV~\cite{caron2020unsupervised} follows a different contrastive approach, where it performs a cluster assignment prediction instead of comparing features of instances directly.
Barlow Twins~\cite{zbontar2021barlow} is a recent method that does not use contrastive learning. It enforces strong correlations between vector representations of distorted versions of the same image, and minimizes the redundancy between components of these vectors. 
Due to their competitive performance, we compare MoCo-V2, SwAV, and Barlow Twins in our experiments and discuss each method in detail in Sec.~\ref{subsec:online_continual_learning_models}.

Self-supervised features are commonly tested by performing a linear evaluation on top of a frozen feature extractor, or by fine-tuning the network on separate tasks. In both cases, the entire dataset is expected to be available for pre-training. \cite{newell2020useful} demonstrates that self-supervision surpasses supervision on several downstream tasks, even when less labeled data is available.
Our work differs from \cite{newell2020useful} in three ways: 1) we test newer self-supervised methods, 2) our downstream task is online continual learning for image classification, which assumes that only a small portion of the data is available for pre-training, and 3) we do not use controlled synthetic datasets and instead focus on high-resolution natural image datasets.

\section{Algorithms}
\label{sec:training_procedure}

In our study, we compare online continual learning systems that use either self-supervised or supervised pre-training as a function of the size of the pre-training dataset. For all experiments, we use ResNet-18 as the CNN. For continual learning with the full-resolution ImageNet dataset, ResNet-18 has been adopted as the universal standard CNN architecture by the community~\cite{rebuffi2017icarl,wu2019large,hayes2019remind,hou2019learning,douillard2020podnet,hayes2019remind,castro2018end, belouadah2019il2m,belouadah2018deesil}. Next, we give details for the pre-training approaches and then we describe the online continual learning algorithms.

\subsection{Pre-Training Approaches}
\label{subsec:pretraining_models}

We describe the four pre-training algorithms we study. The self-supervised methods were chosen based on their strong performance for offline linear evaluation in their papers and because they were practical to train due to their relatively low computational costs.

\textbf{Supervised:} Supervised pre-training for continual learning is our baseline as it is the standard approach used~\cite{hayes2019lifelong,hayes2019remind,rebuffi2017icarl,castro2018end,hou2019learning,wu2019large}, where the CNN is trained using cross-entropy with the labels on the pre-training data. We followed the same protocol as \cite{hayes2019remind} for pre-training, including the use of random resized crop and horizontal flip augmentation from \cite{He_2016_CVPR}. We explore additional augmentations and longer training times in Sec.~\ref{subsec:impact_augswav_epochs}.

\textbf{MoCo-V2:} The original self-supervised MoCo~\cite{he2020momentum} architecture builds a dynamic dictionary such that keys in the dictionary relate to training images. It then trains an encoder network such that query images should be similar to their closest key in the dictionary and further from dissimilar keys using a contrastive loss. MoCo-V2~\cite{chen2020improved} makes two additional improvements over MoCo: it uses an additional projection layer and blur augmentation.

\textbf{Barlow Twins:} Barlow Twins~\cite{zbontar2021barlow} makes the cross-correlation matrix across the outputs of two identical networks fed with distorted versions of an image close to the identity matrix. The objective function enforces the vector representations of distorted versions of the same image to be similar using an \textit{invariance term}, while minimizing the redundancy between the components of these vectors using a \textit{redundancy reduction term}. We chose Barlow Twins due to its novel objective function and competitive performance to contrastive learning methods.

\textbf{SwAV:} The self-supervised SwAV~\cite{caron2020unsupervised} method learns to assign clusters to different augmentations or ``views'' of the same image. Unlike standard contrastive methods, SwAV does not require direct pair-wise feature comparisons for its swapped prediction contrastive loss.

Pre-train models were trained using 4 TitanX GPUs (2015 edition) with 128 GB of RAM. Parameter settings for all approaches are in supplemental materials (Sec. \ref{sec:parameters}.)

\subsection{Online Continual Learning Models}
\label{subsec:online_continual_learning_models}

We evaluate three online continual learning algorithms that use pre-trained features. They were chosen because they have been shown to get strong or state-of-the-art results on incremental class learning for ImageNet. For experiments with class-incremental learning on ImageNet, all continual learning methods first visit the pre-training set with online updates before observing the set that was not used for pre-training. We provide additional details below and parameter settings are in supplemental materials.

\textbf{SLDA}~\cite{hayes2019lifelong}: Deep Streaming Linear Discriminant Analysis (SLDA) keeps the pre-trained CNN features fixed. It solely learns the output layer of the network. It was shown to be extremely effective compared to earlier methods that do update the CNN, despite not using any auxiliary memory. It learns extremely quickly. SLDA stores a set of running mean vectors for each class and a shared covariance matrix, both initialized during pre-training and only these parameters are updated during online training. A new input is classified by assigning it the label of the closest Gaussian in feature space.
    
\textbf{Online Softmax with Replay}: Offline softmax classifiers are often trained with self-supervised features, so we created an online softmax classifier for continual learning by using replay to mitigate forgetting. Like SLDA, it does not learn features in the CNN after pre-training. When a new example is to be learned, it randomly samples a buffer of CNN embeddings to mix in 50 additional samples. This set of 51 samples is then used to update the weights using gradient descent. When the buffer reaches maximum capacity (735K samples = 1.5 GB), we randomly discard an example from the most represented class.
    
\textbf{REMIND}~\cite{hayes2019remind}: Unlike SLDA and Online Softmax with Replay, REMIND does not keep the CNN features fixed after pre-training. Instead, it keeps the lower layers of the CNN fixed, but allows the weights in the upper layers to change. To mitigate forgetting, REMIND replays mid-level CNN features that are compressed using Product Quantization (PQ) and stored in a buffer. After initializing the CNN features during pre-training, we use this same pre-training data to train the PQ model and store the associated compressed representations of this data in the memory buffer. We run REMIND with the settings from \cite{hayes2019remind}.

\section{Experimental Setup}
\label{sec:experiments}

For continual learning, ImageNet ILSVRC-2012 is the standard benchmark for assessing the ability to scale~\cite{russakovsky2015imagenet}. It has 1.2 million images from 1000 classes, with about 1000 images per class used for training. Most existing papers use the first 100 classes (10\%) for supervised pre-training~\cite{rebuffi2017icarl,castro2018end,wu2019large,hayes2019remind} or the first 500 (50\%) classes~\cite{hou2019learning,douillard2020podnet,tao2020topology}. An exception is \cite{hayes2019lifelong}, which studied performance as a function of the size of the supervised pre-training set and found that performance was highly dependent on the amount of data. 

Following previous work \cite{hayes2019remind,hayes2019lifelong}, we randomly select ImageNet classes for pre-training. We split ImageNet into pre-training sets for feature learning of various sizes: 10, 15, 25, 50, 75, 100 classes. In our experiments on ImageNet, after learning features on a pre-train set in an offline manner, the pre-train set and the remainder of the dataset are combined and then examples are fed one-by-one to the learner. Given our limited computational resources, we were not able to train for additional pre-training dataset sizes. In addition to doing continual learning on ImageNet itself, we also study how well the pre-trained features learned on ImageNet from various sizes of pre-train sets generalize to another dataset from an entirely different domain: scene classification. To do this we use the Places-365 dataset~\cite{zhou2017places}, which has 1.8 million images from 365 classes. We only use it for offline linear evaluation and continual learning, i.e., we do not perform pre-training on it. All of our continual learning experiments use the class incremental learning setting, where the data is ordered by class, but all images are shuffled within each class. We report top-1 accuracy on the validation sets.

\section{Results}
\label{sec:results}

\subsection{Offline Linear Evaluation Results}
\label{subsec:offline_results}

We compare the efficacy of self-supervised and supervised features as a function of the amount of pre-train data using the standard \emph{offline} linear evaluation done for self-supervised learning~\cite{he2020momentum,caron2020unsupervised,chen2020simple,zbontar2021barlow}. This lets us measure feature generalization when less data is used, without addressing catastrophic forgetting. Our offline linear evaluation uses a softmax classifier and results are in Table~\ref{tab:linear_eval_results}. Parameter settings are in supplemental materials. 

\input{tables/offline_linear_pre_train}

We compared how effective pre-training approaches were when directly evaluated on the \emph{same} classes used for pre-training. 
We had expected supervised features to outperform self-supervised features in this experiment, but surprisingly SwAV outperformed supervised features when 50 or fewer classes were used and it rivaled or exceeded performance when the number of classes was 75 or 100. MoCo-V2 and Barlow Twins only outperformed supervised features when using 15 or fewer classes.

We assessed the generalization of features to unseen categories by training the softmax classifier on all 1000 classes. SwAV outperformed all other features across all pre-train sizes. The gap in performance between self-supervised and supervised pre-training was larger when using fewer pre-train classes. SwAV outperformed supervised features by 3.24\% and 20.67\% when using 100 and 10 classes during pre-training, respectively.
Thus, self-supervised features generalize better to unseen classes, especially for small pre-train sizes.

We examined the impact of the classes chosen for pre-training by randomly choosing 6 different sets of 10 pre-train classes from ImageNet. We trained an offline softmax classifier using these features on all 1000 ImageNet classes. The mean top-1 accuracy and standard deviation across runs was: 12.65\%$\pm$1.26\% for supervised, 22.78\%$\pm$0.88\% for MoCo-V2, 24.35\%$\pm$0.50\% for Barlow Twins, and 31.88\%$\pm$1.40\% for SwAV. 
With only 10 classes, all self-supervised features outperformed supervised features. 
Since all standard deviations were small, we did not study more seeds for larger pre-train sets.

We evaluated how well the ImageNet pre-trained features transferred to Places-365. Similar to our results on ImageNet, self-supervised features outperformed supervised ones when pre-trained on ImageNet and evaluated on the Places-365 dataset. SwAV features outperformed MoCo-V2 features for all pre-train sizes, where more benefit is seen using fewer pre-train classes, while Barlow Twins outperformed both SwAV and MoCo-V2 when 50 or more classes are used during pre-training. These results demonstrate the versatility of self-supervised features in generalizing to datasets beyond what they were trained on.

\subsection{Online Continual Learning Results}
\label{subsec:online_results}

\input{tables/continual_learning_results}

Table~\ref{tab:main-results} shows online continual learning results with ImageNet for each pre-training method and online learning algorithm. For Deep SLDA, SwAV and Barlow Twins outperform supervised features when the pre-train size is 75 or smaller, but this was only true for MoCo-V2 with a pre-train size of 25 or less. Even when supervised pre-training outperforms self-supervised features, SwAV and Barlow Twins still show competitive performance, with gaps no greater than 4\%. 
For Online Softmax, SwAV performed better than supervised features for all pre-train sizes. Barlow Twins surpasses supervised features for 75 classes or fewer sizes and it is competitive for 100. MoCo-V2 outperforms supervised features for 25 classes or fewer, but it shows competitive performance for other sizes as well. REMIND shows the most benefit from self-supervised pre-training, where SwAV, Barlow Twins, and MoCo-V2 consistently outperform supervised features for all pre-train sizes tested. REMIND using SwAV pre-training on only 10 classes surpasses all results obtained by Deep SLDA. With 50 class pre-training, REMIND with SwAV outperforms supervised pre-training on 100 classes, showing that SwAV can surpass supervised features using half the pre-train data. 

\input{figures/learning_curve_main_paper}

Fig.~\ref{fig:main} shows the relative percentage increase of self-supervision over supervision for REMIND, with a maximum relative improvement of 89.14\% for SwAV, 63.87\% for Barlow Twins, and 60.74\% for MoCo-V2 when only 10 classes are used for pre-training.

Fig.~\ref{fig:learning_curves} shows learning curves for REMIND using SwAV features for various pre-train sizes when multiples of 100 classes have been seen by the model. More learning curves are in supplemental materials.
Adding more classes during pre-training consistently improves REMIND's performance.
To compare with the original REMIND paper, we compute the average top-5 accuracy of REMIND using SwAV features pre-trained on 100 classes of ImageNet and evaluated on seen classes at increments of 100 classes. This variant of REMIND with SwAV achieves an average top-5 accuracy of 83.55\%, which is a 4.87\% absolute percentage improvement over the supervised results in \cite{hayes2019remind}. This variant also achieves a 14.95\% relative increase in final top-1 accuracy over supervised features. This sets a new state of the art result for online learning of ImageNet. 

\paragraph{Domain Transfer with Deep SLDA:}

\input{tables/slda-places}
So far, we have studied the effectiveness of supervised and self-supervised pre-training to generalize to unseen classes from the same dataset. A natural next step is to study how well self-supervised pre-training of features on one dataset generalize to continual learning on another dataset, e.g., pre-train on ImageNet and continually learn on Places-365. This is similar to standard transfer learning and domain adaptation setups~\cite{long2015learning,long2013transfer,pan2010domain,zhang2017joint}, and could prove useful when performing continual learning on small datasets that require generalizable feature representations. To evaluate our hypothesis, we trained Deep SLDA on Places-365 using various pre-trained ImageNet features. We chose Deep SLDA since it is extremely fast to train. We start online learning with the first sample in Places-365 and learn one class at a time from mean vectors initialized as zeros and a covariance matrix initialized as ones, as in \cite{hayes2019lifelong}. Results are in Table~\ref{tab:slda-places}. SwAV features generalized the best for pre-train sizes less than or equal to 25 classes, while Barlow Twins features generalized better for 50 or more pre-train classes. Thus, self-supervised features improve online learning, even when the pre-train and continual datasets differ.

\subsection{Impact of Augmentation and Pre-Training Time on Performance}
\label{subsec:impact_augswav_epochs}

\input{tables/impact_augswav_epochs}

Our main experiments followed \cite{He_2016_CVPR} and used random resized crops and horizontal flip data augmentations for supervised learning. Further, following \cite{hayes2019remind}, we trained supervised models for 40 epochs. However, all self-supervised methods use additional augmentation techniques and longer training times, so we study supervised pre-training under similar conditions.
To do this, we adopt SwAV augmentations (multi-crop, color jitter, gaussian blur, grayscale, and horizontal flips) and training time (400 epochs) for supervised pre-training. We modeled experiments after SwAV since it was the top-performing self-supervised method in Sec.~\ref{subsec:offline_results} and Sec.~\ref{subsec:online_results}. We define two supervised configurations: Supervised Long Training (Supervised$_{LT}$) trains with random resized crop and horizontal flip augmentations~\cite{He_2016_CVPR}; and Supervised Long Training with SwAV Augmentations (Supervised$_{SA}$) trains with SwAV augmentations~\cite{caron2020unsupervised}. Both methods train for 400 epochs with a learning rate of 0.01. All other parameters are the same as in \ref{subsec:parameters_pretraining}. We compare the supervised features with SwAV features in the offline linear setting and online setting using REMIND. The results are in Table~\ref{tab:impact_augswav_epochs}.

Across experiments, Supervised$_{LT}$ features either outperformed or were competitive with supervised features. Similarly, we found that using longer training times and SwAV data augmentations (Supervised$_{SA}$) yielded the best supervised feature performance. These results indicate that longer training and additional data augmentations can improve the quality of supervised features. For offline linear evaluations on the pre-training classes, Supervised$_{SA}$ features always performed better than SwAV features, with a maximum gap of 5.56\%. However, when performing the offline linear evaluation on all 1000 classes, we found that SwAV outperformed or performed competitively with Supervised$_{SA}$. Similarly, SwAV features outperformed all variants of supervised features in the online setting with REMIND. While Supervised$_{LT}$ and Supervised$_{SA}$ features improved supervised performance, SwAV performed competitively or better than supervised features when evaluating on the full dataset.

\section{Discussion and Conclusion}
\label{sec:discussion}

We replaced supervised pre-training with self-supervised techniques for online continual learning on ImageNet. We found that self-supervised methods require fewer pre-training classes to achieve high performance on image classification compared to supervised pre-training. This behavior is seen during offline linear evaluation and online learning and hypothesize that it is due to the generalizability of self-supervised features to unseen classes.

We followed others in using the ResNet-18 CNN architecture~\cite{rebuffi2017icarl,hayes2019lifelong,hayes2019remind}; however, wider and deeper networks have been shown to improve performance, especially for self-supervised learning~\cite{chen2020simple,chen2020big,grill2020bootstrap}. It would be interesting to explore these alternate architectures for online learners initialized with self-supervised features.
Further, since online and batch learning share a similar pre-training procedure, we hypothesize that self-supervised features would benefit incremental batch learning models~\cite{rebuffi2017icarl,wu2019large}.
Future work could also explore self-supervised models pre-trained on large datasets for online continual learning, since self-supervised learning shows even more benefits when trained with large amounts of data.
Moreover, future studies could investigate semi-supervised pre-training. Semi-supervised learning would be advantageous to supervised learning because it requires fewer labels, but could still generalize
similarly to self-supervised techniques. It would also be interesting to develop self-supervised or semi-supervised online updates for continual learning. These could be used to update the plastic layers of REMIND, which could improve performance and generalization further.
Self-supervised features have been evaluated for several downstream tasks in offline settings~\cite{he2020momentum} and it would be interesting to evaluate how well self-supervised features perform for other downstream continual learning tasks, e.g., continual object detection~\cite{acharya2020rodeo,shmelkov2017incremental}, continual semantic segmentation~\cite{cermelli2020modeling,michieli2019incremental}, or continual learning for robotics~\cite{feng2019challenges,lesort2020continual}.
Our work could also facilitate downstream open world learning~\cite{bendale2015towards} or automatic class discovery~\cite{sivic2005discovering,zhu2014unsupervised}, where an agent must identify samples outside of its training distribution as unknown and then learn them. We showed that self-supervised features generalize to unseen classes/datasets in online settings, so extending our work to open world learning would only require a component to identify unknown samples~\cite{liang2018enhancing,hendrycks2017baseline,lee2018simple}.

While much progress has been made to develop continual learners that scale, existing approaches require an initial supervised pre-training phase.
We showed that self-supervised pre-training consistently outperformed supervised pre-training and set a new state-of-the-art for online learning by pairing REMIND with SwAV features. Self-supervised learning is beneficial as it reduces overfitting, promotes generalization, and does not require labels.


\ifthenelse{\boolean{ack}}{
\paragraph{Acknowledgements.}
This work was supported in part by the DARPA/SRI Lifelong Learning Machines program [HR0011-18-C-0051], AFOSR grant [FA9550-18-1-0121], and NSF award \#1909696. The views and conclusions contained herein are those of the authors and should not be interpreted as representing the official policies or endorsements of any sponsor. We thank fellow lab member Manoj Acharya for his comments and useful discussions.
}

\bibliography{egbib}

\ifthenelse{\boolean{combined}}{
\clearpage
\begin{center}
    {\Large Supplemental Material \normalsize}
\end{center}
\input{supplemental-text.tex}

}

\end{document}

%% file: tables/offline_linear_pre_train.tex
\begin{table}[t]
\caption{Top-1 accuracies for offline linear evaluations with different features and varying numbers of pre-train classes. We show top-1 accuracy when evaluating on: 1) only pre-train ImageNet classes, 2) all 1000 ImageNet classes, and 3) all 365 Places classes.\\}
\label{tab:linear_eval_results}
\centering
\footnotesize
\begin{tabular}{clcccccc}
\toprule
\textsc{Evaluation Set} & \textsc{Features} & \textsc{$10$} & \textsc{15} & \textsc{25} & \textsc{50} & \textsc{75} & \textsc{100} \\ 
\midrule
\multirow{4}{*}{\emph{Pre-Train ImageNet}} & Supervised & 71.20 & 82.67 & 86.64 & 83.16 & \textbf{81.20} & 80.32 \\
& MoCo-V2 & 82.20 & 84.00 & 84.40 & 80.80 & 78.96 & 77.76 \\
& Barlow Twins & 85.20 & 85.33 & 83.60 & 78.60 & 75.73 & 74.52 \\
& SwAV & \textbf{91.20} & \textbf{90.27} & \textbf{89.44} & \textbf{85.48} & 81.04 & \textbf{80.82} \\
\midrule
\multirow{4}{*}{\emph{Full ImageNet}} & Supervised & 10.42 & 18.23 & 26.76 & 34.32 & 38.29 & 41.58 \\
& MoCo-V2 & 21.76 & 27.50 & 30.61 & 36.45 & 40.15 & 43.19 \\
& Barlow Twins & 24.90 & 27.79 & 32.46 & 38.63 & 42.16 & 44.66 \\
& SwAV & \textbf{31.09} & \textbf{33.62} & \textbf{35.81} & \textbf{39.54} & \textbf{43.43} & \textbf{44.82} \\
\midrule
\multirow{4}{*}{\emph{Full Places}} & Supervised & 15.07 & 22.83 & 28.46 & 31.03 & 32.63 & 33.78 \\
& MoCo-V2 & 24.07 & 29.45 & 31.01 & 32.58 & 34.54 & 35.93 \\
& Barlow Twins & 29.51 & 31.68 & 33.83 & \textbf{36.94} & \textbf{38.59} & \textbf{39.37} \\
& SwAV & \textbf{32.71} & \textbf{33.88} & \textbf{35.10} & 35.16 & 36.75 & 37.41 \\
\bottomrule
\end{tabular}
\end{table}

%% file: tables/continual_learning_results.tex
\begin{table}[t]
\caption{Final top-1 accuracy values achieved by each continual learning method on ImageNet using various features and numbers of pre-train classes.\\}
\label{tab:main-results}
\centering
\footnotesize
\begin{tabular}{clcccccc}
\toprule
\textsc{Method} & \textsc{Features} & \textsc{10} & \textsc{15} & \textsc{25} & \textsc{50} & \textsc{75} & \textsc{100} \\ 
\midrule
\multirow{4}{*}{\emph{Deep SLDA}} & Supervised & 9.53 & 14.67 & 20.7 & 26.54 & 29.53 &  \textbf{31.99} \\
& MoCo-V2 & 19.37 & 20.66 & 21.64 & 24.43 & 26.26 & 28.31 \\
& Barlow Twins & 18.48 & 20.10 & 23.02 & 27.45 & 30.25 & 31.81 \\
& SwAV &  \textbf{22.33} &  \textbf{24.01} &  \textbf{25.22} &  \textbf{28.5} &  \textbf{30.89} & 31.77 \\
\midrule
\multirow{4}{*}{\emph{Online Softmax}} & Supervised & 9.87 & 15.92 & 23.58 & 30.16 & 33.05 & 35.14 \\
& MoCo-V2 &  \textbf{23.09} &  \textbf{25.67} & 25.66 & 28.79 & 31.14 & 33.81 \\
& Barlow Twins & 20.30 & 22.18 & 26.01 & 30.47 & 33.39 & 35.06 \\
& SwAV & 18.65 & 21.40 &  \textbf{27.42} &  \textbf{35.79} &  \textbf{39.48} &  \textbf{41.31} \\
\midrule
\multirow{4}{*}{\emph{REMIND}} & Supervised & 19.79 & 26.17 & 33.48 & 39.38 & 43.40 & 45.28 \\
& MoCo-V2 & 31.81 & 34.39 & 38.43 & 44.50 & 47.36 & 49.25 \\
& Barlow Twins & 32.43 & 34.52 & 37.94 & 42.65 & 45.30 & 46.85 \\
& SwAV & \textbf{37.43} & \textbf{40.09} & \textbf{43.05} & \textbf{48.31} & \textbf{51.01} & \textbf{52.05} \\
\bottomrule
\end{tabular}
\end{table}

%% file: figures/learning_curve_main_paper.tex
\begin{wrapfigure}[16]{t}{0.35\textwidth}
 \centering
    \includegraphics[width=\linewidth]{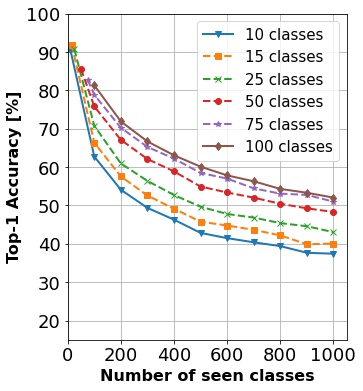}
    \vspace{0.5pt}
    \caption{Learning curve for REMIND with SwAV features at various pre-train sizes.}
\label{fig:learning_curves}
\end{wrapfigure}

%% file: tables/slda-places.tex
\begin{table}[t]
\caption{Final top-1 accuracies for Deep SLDA when pre-trained on ImageNet using various features and pre-train classes and continually updated and evaluated on Places-365.\\}
\label{tab:slda-places}
\centering
\footnotesize
\begin{tabular}{lcccccc}
\toprule
\textsc{Features} & \textsc{10} & \textsc{15} & \textsc{25} & \textsc{50} & \textsc{75} & \textsc{100} \\ 
\midrule
Supervised & 13.77 & 19.59 & 23.42 & 25.28 & 26.04 & 27.03 \\
MoCo-V2 & 23.36 & 23.72 & 24.05 & 24.17 & 25.11 & 26.00 \\
Barlow Twins & 24.14 & 25.63 & 27.27 & \textbf{29.21} & \textbf{30.27} & \textbf{30.90} \\
SwAV & \textbf{26.53} & \textbf{27.20} & \textbf{28.22} & 28.48 & 29.38 & 29.58 \\
\bottomrule
\end{tabular}
\end{table}

%% file: tables/impact_augswav_epochs.tex
\begin{table}[t]
\caption{Final top-1 accuracies on ImageNet for offline linear evaluations and online learning using REMIND using various pre-train sizes. We compare the performance Supervised$_{LT}$, Supervised$_{SA}$, Supervised, and SwAV features. For offline linear, we evaluate features on both the pre-train classes and all 1000 classes. REMIND evaluates on all 1000 classes.\\
}
\label{tab:impact_augswav_epochs}
\centering
\footnotesize
\begin{tabular}{ccccccccc}
\toprule
\textsc{Method} & \textsc{Eval. Set} & \textsc{Features} & \textsc{10} & \textsc{15} & \textsc{25} & \textsc{50} & \textsc{75} & \textsc{100} \\ 
\midrule
\multirow{4}{*}{\emph{Offline Linear}} &\multirow{4}{*}{\emph{Pre-Train}} & Supervised & 71.20 & 82.67 & 86.64 & 83.16 & 81.20 & 80.32 \\
&& Supervised$_{LT}$ & 91.20 & 87.87 & 90.08 & 85.52 & 81.63 & 80.92\\
&& Supervised$_{SA}$ & \textbf{94.80} & \textbf{94.00} & \textbf{91.92} & \textbf{88.48} & \textbf{86.56} & \textbf{85.66} \\
&& SwAV & 91.20 & 90.27 & 89.44 & 85.48 & 81.04 & 80.82 \\
\midrule
\multirow{4}{*}{\emph{Offline Linear}} &\multirow{4}{*}{\emph{Full}} & Supervised & 10.42 & 18.23 & 26.76 & 34.32 & 38.29 & 41.58 \\
&& Supervised$_{LT}$ & 24.85 & 27.34 & 30.89 & 34.11 & 36.37 & 39.80 \\
&& Supervised$_{SA}$ & 27.88 & 31.34 & 35.13 & \textbf{39.57} & 42.28 & 44.76 \\
&& SwAV & \textbf{31.09} & \textbf{33.62} & \textbf{35.81} & 39.54 & \textbf{43.43} & \textbf{44.82} \\
\midrule
\multirow{4}{*}{\emph{Online (REMIND)}} &\multirow{4}{*}{\emph{Full}} & Supervised & 19.79 & 26.17 & 33.48 & 39.38 & 43.40 & 45.28 \\
&& Supervised$_{LT}$ & 31.49 & 33.71 & 36.64 & 41.07 & 43.59 & 45.88 \\
&& Supervised$_{SA}$ & 34.59 & 37.35 & 42.23 & 46.20 & 49.32 & 50.82 \\
&& SwAV & \textbf{37.43} & \textbf{40.09} & \textbf{43.05} & \textbf{48.31} & \textbf{51.01} & \textbf{52.05} \\
\bottomrule
\end{tabular}
\end{table}

%% file: supplemental-text.tex
\beginsupplement

\section{Parameter Settings}
\label{sec:parameters}

\subsection{Pre-Training Approaches}
\label{subsec:parameters_pretraining}

We provide parameter settings for each pre-training approach below.

\begin{itemize}[noitemsep, nolistsep]
    \item \textbf{Supervised:}
    Following~\cite{hayes2019remind}, systems were trained for 40 epochs with a minibatch size of 256, an initial learning rate of 0.1 with a decrease of 10$\times$ every 15 epochs, momentum of 0.9, and weight decay of 1e-4. Standard random crop and horizontal flips were used for augmentation, unless noted otherwise. The network with best performance on the validation set is picked.
 
    \item \textbf{MoCo-V2:}
    Following \cite{chen2020improved}, we train MoCo-V2 models for 800 epochs using a learning rate of 0.015 and minibatch size of 128. 

    \item \textbf{Barlow Twins:}
    Following \cite{zbontar2021barlow}, we train Barlow Twins models for 300 epochs with a minibatch size of 1024, learning rate of 0.2, and a trade-off parameter $\lambda$ of 0.0051.
    
    \item \textbf{SwAV:}
    Following \cite{caron2020unsupervised}, SwAV models are trained for 400 epochs with an initial learning rate of 0.6, final learning rate of 0.0006, epsilon of 0.03, and minibatch size of 64. We set the number of prototypes to $10 \left| \mathcal{Y} \right|$ in a queue of length 384.

\end{itemize}

\subsection{Offline Linear Evaluations}
\label{subsec:parameters_offline}

To linearly evaluate the pre-trained features on the ImageNet dataset, the classifier was trained for 100 epochs with a minibatch size of 256. For supervised features with SwAV and Barlow Twins, we used a learning rate of 0.1 which we decay by a factor of 10 at 60 and 80 epochs and L2 weight decay of 1e-5. These settings did not work well for MoCo-V2, so we used the settings for linear evaluation from \cite{he2020momentum} instead, which were a learning rate of 30 that we decay by a factor of 10 at epochs 60 and 80 with no weight decay.

For linear evaluation on the Places-365 dataset, we trained the softmax classifier for 28 epochs with a minibatch size of 256. We used the same learning rates and L2 weight decays from the ImageNet linear evaluation and reduced the learning rate by 10$\times$ at epochs 10 and 18. 

\subsection{Online Continual Learning Methods}
\label{subsec:parameters_online}

We provide parameter settings for each continual learning method below.
\begin{itemize}[noitemsep, nolistsep]
    \item \textbf{SLDA:}
    Following \cite{hayes2019lifelong}, we use a plastic covariance matrix and shrinkage of 1e-4. 
    
    \item \textbf{Online Softmax with Replay:}
    We use a learning rate of 0.1. The buffer contains a maximum of 735K feature vectors each with 512 dimensions (1.5 GB). This buffer size was chosen to match the size of the buffer for REMIND in GB.
    
    \item \textbf{REMIND:}
    We follow the parameter settings from \cite{hayes2019remind}, i.e., starting learning rate of 0.1, 32 codebooks each of size 256, 50 randomly selected replay samples, a buffer size of 959,665 (equal to 1.5 GB), manifold mixup and random resize crop data augmentation, and extracting mid-level features such that two convolutional layers and the final classification layer remain plastic during online learning.

\end{itemize}
\section{Additional Results}

\subsection{Relative Performance Improvements}

\begin{figure*}[h]
     \centering
     \begin{center}
     \begin{subfigure}[b]{0.48\textwidth}
         \centering
         \includegraphics[width=\textwidth]{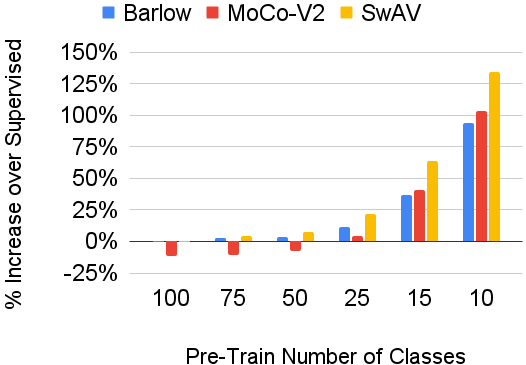}
         \caption{Deep SLDA}
         \label{subfig:relative_perf_slda}
     \end{subfigure}
     \hfill
     \begin{subfigure}[b]{0.48\textwidth}
         \centering
         \includegraphics[width=\textwidth]{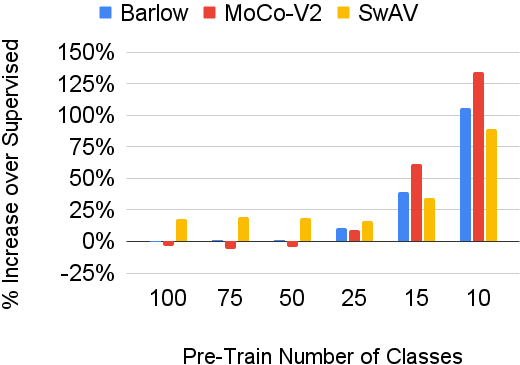}
         \caption{Online Softmax}
         \label{subfig:relative_perf_softmax}
     \end{subfigure}
     \end{center}
        \caption{Relative performance improvements over the supervised baseline using MoCo-V2, Barlow Twins, and SwAV features for online continual learning with (a) Deep SLDA and (b) Online Softmax on ImageNet.}
        \label{fig:relative_perf_additional}
        \vspace{3 pt}
\end{figure*}

Fig.~\ref{fig:relative_perf_additional} shows the relative performance improvements exhibited by the Deep SLDA and Online Softmax methods when performing online continual learning on the ImageNet dataset using MoCo-V2, Barlow Twins, and SwAV features. 

Deep SLDA (\ref{subfig:relative_perf_slda}) shows a maximum relative improvement of 134.31\% for SwAV features, 93.91\% for Barlow Twins features, and 103.25\% for MoCo-V2 features when only 10 classes are used for pre-training. Deep SLDA follows the same trend as REMIND from Fig.~\ref{fig:main}, with SwAV outperforming Barlow Twins and MoCo-V2 for all pre-train sizes. MoCo-V2 shows negative relative improvement for 50, 75, and 100 pre-training classes, while SwAV and Barlow Twins show small negative relative improvements of -0.69\% and -0.56\% respectively for 100 classes. These few cases of negative relative performances occur when there are a large number of pre-training classes, which is less desirable for pre-training as it requires more data.

Online Softmax (\ref{subfig:relative_perf_softmax}) shows a maximum relative improvement of 88.96\% for SwAV features, 105.67\% for Barlow Twins features, and 133.94\% for MoCo-V2 features when only 10 classes are used for pre-training. Surprisingly, MoCo-V2 outperforms SwAV and Barlow Twins for 10 and 15 pre-train classes, which differs from the results for REMIND and Deep SLDA. We see a small negative relative improvement with MoCo-V2 for 50, 75 and 100 pre-train classes, while Barlow Twins shows a small negative relative improvement for 100 pre-train classes only. SwAV shows positive relative improvement for all pre-train sizes. Overall, these results demonstrate that self-supervised features are superior to supervised features for continual learning when less data is used during pre-training.

\subsection{Learning Curves}

Learning curves for online continual learning on ImageNet using REMIND, Deep SLDA and Online Softmax with various features are in Fig.~\ref{fig:learning_curves_remind}, Fig.~\ref{fig:learning_curves_slda} and Fig.~\ref{fig:learning_curves_softmax}, respectively. These curves show the top-1 accuracy every time a multiple of 100 classes has been seen by the model, including the performance right after pre-training.

Learning curves for REMIND using supervised, MoCo-V2, Barlow Twins, and SwAV features for various pre-training set sizes show that adding more classes during pre-training consistently improves REMIND's performance for all features used. Furthermore, using SwAV, Barlow Twins, or MoCo-V2 consistently improves performance over supervised pre-training, which can be seen by the vertical shift of the learning curves across different features.

Similar to REMIND, adding more classes during pre-training improves Deep SLDA's performance for all features used. The vertical shift of the curves upwards across features shows that MoCo-V2, Barlow Twins, and SwAV consistently outperform supervised features for 10, 15 and 25 pre-train classes. Even though supervised features show the best performance for 100 classes, MoCo-V2, Barlow Twins, and SwAV show competitive results. 

Online Softmax also has better performance when using more pre-train classes across different features. Surprisingly, MoCo-V2 outperforms Barlow Twins, SwAV, and supervised features for 10 and 15 pre-train classes. SwAV features outperform supervised features across all pre-train sizes. It is worth noting that when 100 pre-train classes are used, MoCo-V2, SwAV, and supervised features start around the same top-1 accuracy, but after the model has finished learning all 1000 classes, SwAV achieves the highest performance. This behaviour means that SwAV features obtained during pre-training on 100 classes are more useful when continually learning all 1000 classes.

\input{figures/learning_curve_figure_remind}

\input{figures/learning_curve_figure_slda}

\input{figures/learning_curve_figure_softmax}

%% file: figures/learning_curve_figure_remind.tex
\begin{figure*}[t]
     \centering
     \begin{center}
     \begin{subfigure}[b]{0.47\textwidth}
         \centering
         \includegraphics[width=\textwidth]{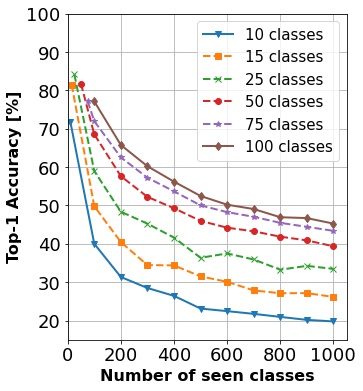}
         \caption{Supervised}
         \label{subfig:supervised-lc}
     \end{subfigure}
     \hfill
     \begin{subfigure}[b]{0.47\textwidth}
         \centering
         \includegraphics[width=\textwidth]{images/REMIND_learning_curves_swav.jpg}
         \caption{SwAV}
         \label{subfig:swav-lc}
     \end{subfigure}\\
     
     \begin{subfigure}[b]{0.47\textwidth}
         \centering
         \includegraphics[width=\textwidth]{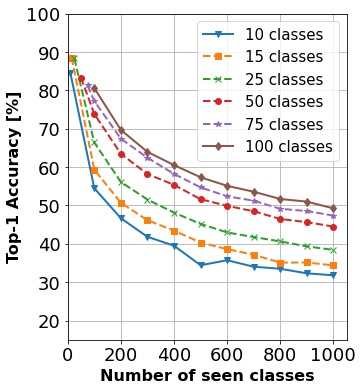}
         \caption{MoCo-V2}
         \label{subfig:mocov2-lc}
     \end{subfigure}
     \hfill
     \begin{subfigure}[b]{0.47\textwidth}
         \centering
         \includegraphics[width=\textwidth]{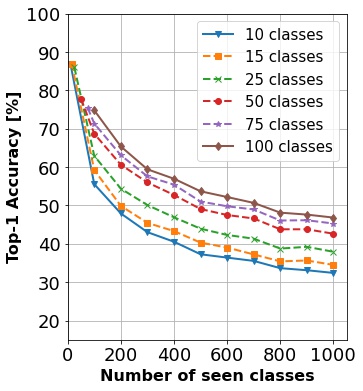}
         \caption{Barlow Twins}
         \label{subfig:barlow-lc}
     \end{subfigure}
     \end{center}

        \caption{Learning curves on ImageNet for each pre-train size with REMIND using (a) supervised, (b) SwAV, (c) MoCo-V2, and (d) Barlow Twins features.}
        \label{fig:learning_curves_remind}
\end{figure*}

%% file: figures/learning_curve_figure_slda.tex
\begin{figure*}[t]
     \centering
     \begin{center}
     \begin{subfigure}[b]{0.47\textwidth}
         \centering
         \includegraphics[width=\textwidth]{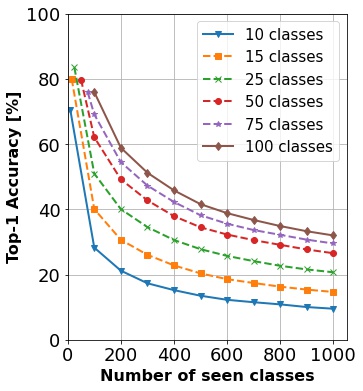}
         \caption{Supervised}
         \label{subfig:supervised-lc-slda}
     \end{subfigure}
     \hfill
     \begin{subfigure}[b]{0.47\textwidth}
         \centering
         \includegraphics[width=\textwidth]{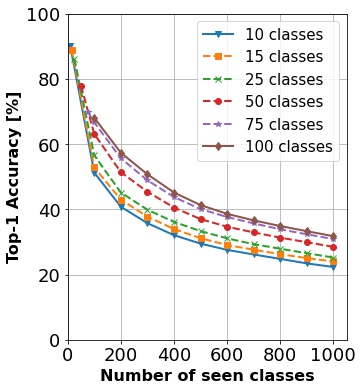}
         \caption{SwAV}
         \label{subfig:swav-lc-slda}
     \end{subfigure}\\
     
     \begin{subfigure}[b]{0.47\textwidth}
         \centering
         \includegraphics[width=\textwidth]{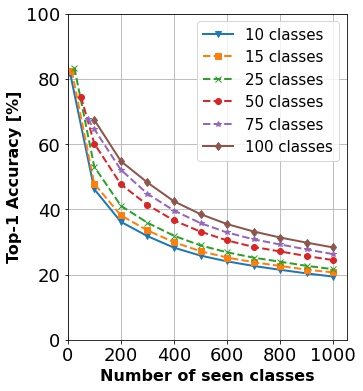}
         \caption{MoCo-V2}
         \label{subfig:mocov2-lc-slda}
     \end{subfigure}
     \hfill
     \begin{subfigure}[b]{0.47\textwidth}
         \centering
         \includegraphics[width=\textwidth]{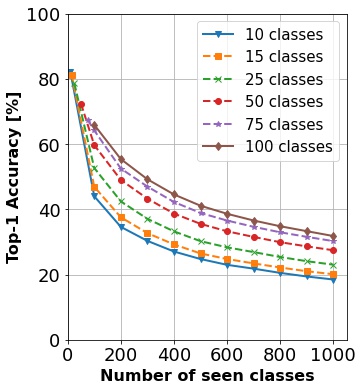}
         \caption{Barlow Twins}
         \label{subfig:barlow-lc-slda}
     \end{subfigure}
     \end{center}
        \caption{Learning curves on ImageNet for each pre-train size with Deep SLDA using (a) supervised, (b) SwAV, (c) MoCo-V2, and (d) Barlow Twins features.}
        \label{fig:learning_curves_slda}
\end{figure*}

%% file: figures/learning_curve_figure_softmax.tex
\begin{figure*}[t]
     \centering
     \begin{center}
     \begin{subfigure}[b]{0.47\textwidth}
         \centering
         \includegraphics[width=\textwidth]{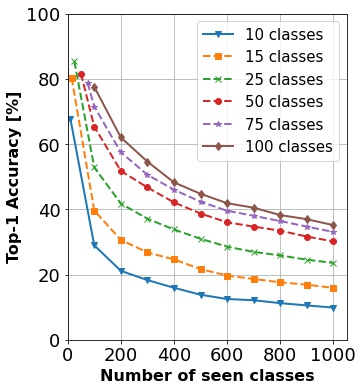}
         \caption{Supervised}
         \label{subfig:supervised-lc-softmax}
     \end{subfigure}
     \hfill
     \begin{subfigure}[b]{0.47\textwidth}
         \centering
         \includegraphics[width=\textwidth]{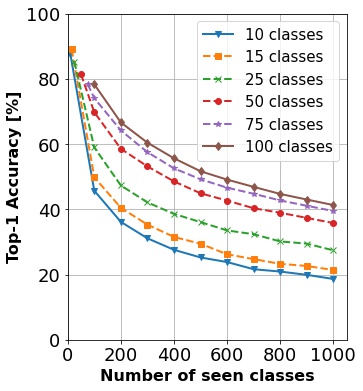}
         \caption{SwAV}
         \label{subfig:swav-lc-softmax}
     \end{subfigure}\\
     
     \begin{subfigure}[b]{0.47\textwidth}
         \centering
         \includegraphics[width=\textwidth]{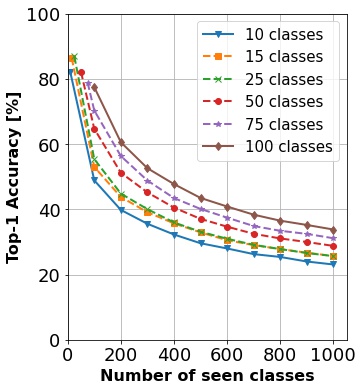}
         \caption{MoCo-V2}
         \label{subfig:mocov2-lc-softmax}
     \end{subfigure}
     \hfill
     \begin{subfigure}[b]{0.47\textwidth}
         \centering
         \includegraphics[width=\textwidth]{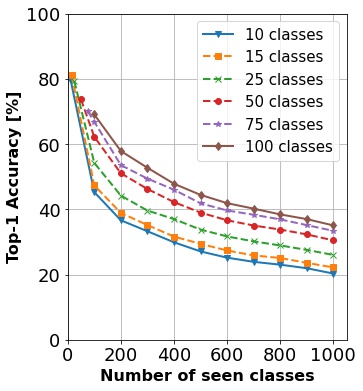}
         \caption{Barlow Twins}
         \label{subfig:barlow-lc-softmax}
     \end{subfigure}
     \end{center}
        \caption{Learning curves on ImageNet for each pre-train size with Online Softmax using (a) supervised, (b) SwAV, (c) MoCo-V2, and (d) Barlow Twins features.}
        \label{fig:learning_curves_softmax}
\end{figure*}